\documentclass[runningheads]{llncs}
\usepackage[pdftex]{graphicx}
\usepackage[pdftex]{color}
\usepackage{amsmath,amssymb} 
\usepackage{comment}
\usepackage{multirow}
\usepackage{here}

\begin{document}
\pagestyle{headings}
\mainmatter

\def\ACCV20SubNumber{815}  

\title{MCGKT-Net: Multi-level Context Gating Knowledge Transfer Network for Single Image Deraining} 
\titlerunning{MCGKT-Net for Single Image Deraining}

%
\author{Kohei Yamamichi\orcidID{0000-0001-7304-6438} and Xian-Hua Han\orcidID{0000-0002-5003-3180}}
\authorrunning{K. Yamamichi et al.}
%

\institute{Graduate School of Science and Technology for Innovation, Yamaguchi University, 1677-1 Yoshida, Yamaguchi City, Yamaguchi, 753-8511, Japan.\\
\email{\{a035vbu,hanxhua\}@yamaguchi-u.ac.jp}}

\maketitle

\begin{abstract}
Rain streak removal in a single image is a very challenging task due to its ill-posed nature in essence. Recently, the end-to-end learning techniques with deep convolutional neural networks (DCNN) have made great progress in this task. However, the conventional DCNN-based deraining methods have struggled to exploit deeper and more complex network architectures for pursuing better performance. This study proposes a novel MCGKT-Net for boosting deraining performance, which is a naturally multi-scale learning framework being capable of exploring multi-scale attributes of rain streaks and different semantic structures of the clear images. In order to obtain high representative features inside MCGKT-Net, we explore internal knowledge transfer module using ConvLSTM unit for conducting interaction learning between different layers and investigate external knowledge transfer module for leveraging the knowledge already learned in other task domains. Furthermore, to dynamically select useful features in learning procedure, we propose a multi-scale context gating module in the MCGKT-Net using squeeze-and-excitation block. Experiments on three benchmark datasets: Rain100H, Rain100L, and Rain800, manifest impressive performance compared with state-of-the-art methods.
\end{abstract}


\section{Introduction}
With the rapid development of image acquisition technology, it has become possible to obtain more and more high-quality images than ever before and has witnessed major advances in various practical computer vision-based systems. However, the images captured under adverse weather conditions such as rain, haze, night have been greatly degraded and significantly affect the performance of the vision systems such as surveillance and autonomous driving \cite{App}. Therefore, automatically recovering the underlying clean single rainy image, simply named as image deraining.

Traditional image deraining methods mainly exploit simple linear-mapping transformation motivated by the physical rainy model that the observed rainy image: {\bf O} can be generally modeled as a linear sum of a rain-free clean image: {\bf B} and the rain streak: {\bf R}  \cite{Rain1,Rain2}. The linear model as mathematical representation can be formulated as:
\begin{equation}
{\bf O}={\bf B}+{\bf R}             
\end{equation}

Given Eq. (1), the deraining method aims at removing {\bf R} from {\bf O} to get {\bf B}. Since the number of the unknown variables in {\bf B} and {\bf R} are 2 times of the known ones in {\bf O} and there are many solutions of {\bf B}, {\bf R} for a given {\bf O}, this is naturally an ill-posed problem. For well solving this problem, previous methods  \cite{Opt1,Opt2,Opt3} mainly focus on employing various hand-crafted priors for exploring the underlying structure of the latent clean image or the attribute of the rain streaks. However, to hammer out the proper prior for a specific image and a kind of rain streak remains to be an art.  
Recent approaches for single image deraining leverage deep learning to mitigate the dependence on the hand-crafted priors, and illustrate that the convolutional neural network (CNN \cite{CNN,Du1,Wang}) itself can effectively capture the intrinsic characteristics of the latent clean images via learning strategy. In the deep learning-based scenario for single image deraining, various CNN models have been proposed and revolved to more and more complicated architectures for pursuing high performance. Although the deep learning-based approaches manifest significant improvement in single image deraining, there still exist some limitations. As mentioned above, the CNN models have progressed into much more complex and diverse architectures to boost the performance, and thus lead to difficulty for practical implementation and model training. In addition, most existing CNN models adopt a single-scale framework for feature representation, which rarely captures the underlying correlation of rain streaks across scales. Recently, authors in \cite{Zheng},\cite{Jiang} exploited a multi-scale deep framework for image deraining. Unfortunately, these exploitations fail to make full use of the interactive correlation of multi-scale rain streaks and have complicated architectures with several subnets. Furthermore, the exiting CNN methods generally get to start training with the training pairs of the observed rainy images and their corresponding clean images from scratch and cannot exploit the knowledge existed in the already learned CNN models from the clean training images in other task domains such as image classification.

To handle with the above limitations, we propose a novel deraining network, called as multi-level context-gating knowledge transfer network: MCGKT-Net. The MCGKT-Net is based on the well-known U-Net architecture, which can be simply implemented and naturally a multi-scale learning framework. In order to exploit the correlation of the low-level features and high-level features in multi-scale encoder and decoder subnets of U-Net \cite{U-Net}, we employ a ConvLSTM unit for interactively transferring the learned knowledge of two sides instead of directly duplicating the encoder’s feature to  the decoder side, called as internal knowledge transfer. Since training CNN model with rain-degraded images as input possibly leads to model deviation from optimal parameters, we advocate to transfer a part of knowledge (the shallow layer's parameters) hold in the learned CNN model with the clean training images in other task domains such as image classification, and reuse them in our proposed MCGKT-Net for boosting deraining performance, called as external knowledge transfer. Finally, we adopt squeeze-and-excitation block in multi-scale learned features of the decoder subnet for dynamically selecting the useful learned contexts for being inputted to the next scale, called as multi-scale context gating. Experiments on three benchmark datasets demonstrate the promising performance compared with the state-of-the-art methods on image deraining. 

In summary, our main contributions are three-fold:
\begin{enumerate}
\item We present a simply-implemented and naturally multi-scale deraining network, which can effectively explore the multi-scale attributes of rain streak and different underlying semantic structures of the clean images;
\item We exploit the interactive learning between the same level features of encoder and decoder subnets for internal knowledge transfer and reuses the existed knowledge learned in other task domains for external knowledge transfer to boosting deraining performance;
\item We explore a multi-scale context gating module for dynamically selecting useful features of decoder subnet using squeeze-and-excitation block.    
\end{enumerate}  

The rest of this paper is organized as follows. Section 2 surveys the related work including deep learning-based image deraining methods and multi-level learning networks. Section 3 presents the proposed MCGKT network for image deraining. Extensive experiments are conducted in Sec. 4 to compare the proposed MCGKT-Net with state-of-the-art image deraining methods on three benchmark datasets. The conclusion is given in Sec. 5.


\section{Related Work}
In the past decades, image deraining has been actively researched in the low-level computer vision community, and substantial improvements have been witnessed. This work mainly concentrates on the more challenge deraining from a single image. Here, we briefly survey the related work. 

\subsection{Single Image Deraining}

Rain streaks removal from a single image is an extremely challenging task due to its ill-posed nature. Previous methods are mainly divided into two categories: optimization-based methods \cite{Chen1,Kang,Chang,Ren1} and deep learning-based methods \cite{Fu1,DDN,DIDMDN,Zhang2}. Optimization based methods usually formulate the deraining task as a mathematical model motivated by the fact that rainy images can be decomposed into a clean background image layer and a rain layer. To recover more robust clean image, the prior knowledge for characterizing the underlying structure of the latent clean image layer and the attribute of the rain layers has imposed on the formulated mathematical model as regularization term and employed optimization strategy for solving. Kang et al. \cite{Kang} proposed to apply sparse coding to separate rain streaks from the high-frequency layer, while Luo et al. \cite{Luo} explored a discriminative sparse coding framework for modeling image patches. The work by Chen et al. \cite{Chen1} and Chang et al. \cite{Chang} leveraged the low-rank property of rain streaks for removing the decomposed rain layer based on low-rankness. Since the composite models \cite{Chen1,Kang,Chang} regularized by modeling the sparse and low-rank prior are insufficient in characterizing the decomposed layers, leading to limited deraining performance on diverse images. In addition, the explored priors (for example sparsity, low-rankness) on the previous approaches are hand-crafted, and to discover a proper prior for a specific image and a kind of rain type still remains be an art or requires comprehensive analysis for a specific rainy image. 
Recently, deep convolutional neural network has been widely applied to single image deraining, and validated that promising performance can be achieved \cite{Fu1,DDN,DIDMDN,Zhang2}. Fu et al. \cite{Fu1} first explored a three-layer convolutional network to predict clean image high-frequency component from its rain-contaminated counterpart, and further extended it to a 20-layer CNN structure by incorporating Residual-Block, called as deep detail network, for pursuing better performance \cite{DDN}. Zhang et al. \cite{DIDMDN} presented a multi-stream dense network for joint rain density estimation and deraining. To generate more visually plausible deraining result, the same research group investigated a conditional generative adversarial network (GAN) for single image deraining \cite{Zhang2}, and proved to achieve visually high-quality reconstructions. In \cite{RESCAN}, a novel deep network architecture based on recurrent neural networks and squeeze-and-excitation context aggregation module (RESCAN) has been proposed and adaptively adjusted parameters for various rain streak layers. Fan et al. \cite{Fan} proposed residual-guide network with recursive convolution module and multi-level supervision not only on the final results but also on the intermediate results progressively for predicting high-quality reconstruction. Wei et al. \cite{SEMI} proposed a semi-supervised image deraining network, while Ren et al. \cite{PReNet} focused on several factors including network architecture for integrating progressive ResNet and recurrent layers inside and cross stages, and loss functions, and provided a better and simpler baseline deraining network.

\subsection{Multi-level learning network}

It is known that a rainy image is possibly decomposed into the image layer and rain streak layer, which may consist of multiple layers, especially under heavy rain conditions. The rain streaks decomposed in multiple layers manifest intrinsically multi-scale attributes and some self-similarity properties within and across scales, which is prospected to boost the deraining performance via exploring the correlated information between and across multiple levels of rain layers. Most existing methods recur to deeper and more complex network architecture for pursuing better deraining performance but cannot make full use of the underlying correlation between and across different rain layers. Although, a few work \cite{LPNet,Zheng,Jiang} have been investigated to explore multiscale information for deraining from a single image, which mainly leverages multiple subnets (several mainstreams) for exploiting different scales, and lead to more complicated DCNN architectures. 
As we know that the convolutional encoder-decoder network itself is a multi-level learning architecture, where the encoder path learns feature representation evolved to large scale with the increased depth of the network while the decoder path attempts to recover the feature representation with more detailed structure (small scale) from the final output of the encoder with more semantic information and large context. Further, to retain more detail structures in the final prediction results, skip connection is usually used for duplicating the feature representation of the encoder path to the decoder side in the same level such as in U-Net \cite{U-Net}, FCN \cite{FCN}.  Although the encoder-decoder network has also been adopted for image deraining \cite{Du1,Du2,Wang}, most existing methods cannot effectively exploit the correlation of the feature representations and compensate each other between the encoder and decoder paths. This study explores a novel and simple deraining network, called multi-level context gating knowledge transfer network (MCGKT-Net), which is based on the multi-level encoder-decoder network and investigates both internal and external knowledge transfer for boosting deraining performance.

\section{The proposed MCGKT-Net}
The mainstream of our proposed MCGKT-Net follows the encoder-decoder network architecture, and multi-level feature representation can be learned in both encoder and decoder paths. To effectively explore the feature interaction and correlation between encoder and decoder paths, we propose to leverage backward ConvLSTM blocks instead of simple duplication, to transfer the semantic structure of the high-level features in the decoder path to the encoder side, and input the interactively learned features to the subsequent level of the decoder, called as internal knowledge transfer module (IKT). We further leverage a part of knowledge (the shallow layer's parameters) maintained in the learned CNN model with the clean training images in other task domains such as image classification for training the deraining model from a good initial state, called as external knowledge transfer module (EKT). In addition, to adaptively select more useful feature representations in the learning procedure, we exploit squeeze and excitation block (SE) to the multi-level features in the decoder path for constructing the multi-level context gating module (MLCG). Then the MCGKT-Net consists of the mainstream of the encoder-decoder architecture, the knowledge transfer module with IKT and EKT, and the MLCG module. The schematic concept of the proposed MCGKT-Net is shown in Fig. 1. Next, we would describe the different parts of the MCGKT-Net.

\subsection{The mainstream of the encoder-decoder architecture}

The mainstream of our used network architecture consists of two paths: encoder and decoder, and each path is divided into four blocks. Both encoder and decoder paths learn multi-level feature representations in the multiple blocks, where encoder employs MaxPooling layer with a 2*2 kernel for decreasing feature map size to half in both horizontal and vertical directions between blocks while decoder performs up-sampling for doubly recovering the feature map size between blocks. In each block of both encode and decoder, we implement it in 3 convolutional layers with 3*3 kernels following ReLU activation function. The channel number of the learned feature maps block-wisely is doubled in the encoder while is halve in the decoder. Thus the architecture in the encoder path is same as the first 3 shallow layers of the popularly used VGGNet in different vision problems.

\begin{figure}[h]
\centering
\includegraphics[width=\textwidth]{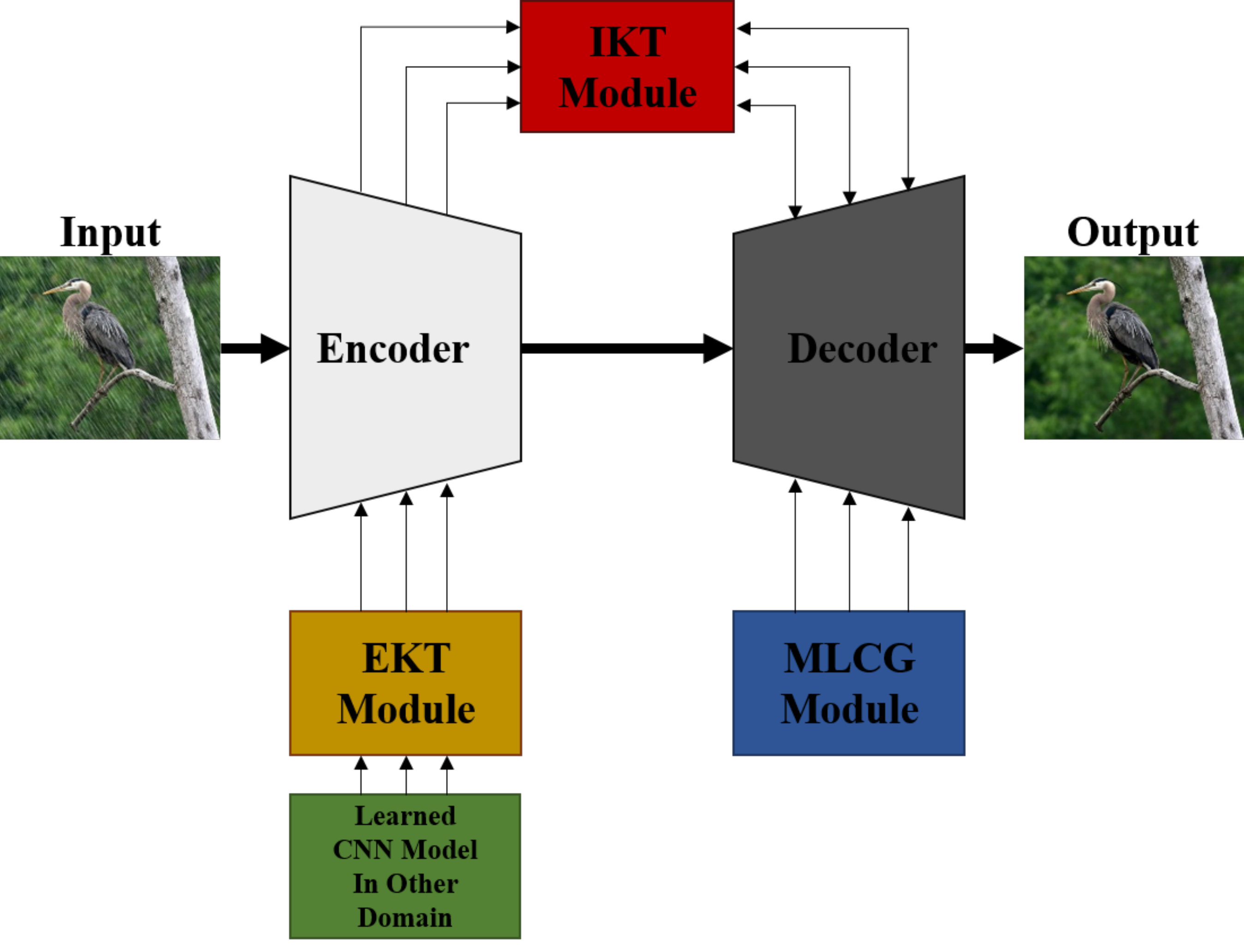} 
\caption{Network architecture of MCGKT-Net. The abbreviations IKT, EKT and MLCG Module denote internal knowledge transfer, external knowledge transfer module, and multi-level context gating module, respectively.}
\label{fig:network_architecture}
\end{figure}

Let's denotes the input and output of the $j-th$ block in the encoder path as ${\bf E}_{i}^{j}$, ${\bf E}_{o}^{j}$, and in the decoder path as ${\bf D}_{i}^{j}$, ${\bf D}_{o}^{j}$, respectively, the relation of the input and output of each block can be expressed as:
\begin{equation}
{\bf E}_{o}^{j}=f({\bf E}_{o}^{j}, \theta_{E,i}) , \; {\bf D}_{o}^{j}=f({\bf D}_{o}^{j}, \theta_{D,i})
\end{equation}
where $f(\cdot)$ represents the transformation operation of 3 convolutional layers with the learned parameters $\theta_{E,i}$ and $\theta_{D,i}$, respectively. The $j+1-th$ inputs in the encoder and the $j-1-th$ output in the decoder sides can be obtained from the $j-th$ outputs, and are expressed as:
\begin{equation}
{\bf E}_{i}^{j+1}=MP({\bf E}_{o}^{j}), \; {\bf D}_{i}^{j-1}=UP({\bf D}_{o}^{j})
\end{equation}
where $ MP(\cdot)$ and $UP(\cdot)$ denote the MaxPooling and up-sampling layer implemented between blocks, respectively. Then the mainstream of our encoder-decoder network can learn multi-level feature representations: ${\bf E}_{i}=[{\bf E}_{i}^{1}, {\bf E}_{i}^{2}, {\bf E}_{i}^{3}, {\bf E}_{i}^{4}]$ in the encoder and ${\bf D}_{o}=[{\bf D}_{o}^{1}, {\bf D}_{o}^{2}, {\bf D}_{o}^{3}, {\bf D}_{o}^{4}]$ in the decoder with different context information, where the encoder generates low-level features with more detail image structure while decoder results in high-level feature representation with more semantic information. The encoder-decoder architecture is very simple with the convolutional layer of small kernel size 3*3 and is prospected to be easily trained in different vision tasks. In this network, it is known that the outputs of the deeper blocks inherit from the shallow blocks of the decoder and all blocks of the encoder, and produce more semantic information but may lose detail structure. To maintain more detail structure in the final results, the existing encoder-decoder architecture in different vision tasks such as FCN, U-Net usually employ skip connection for directly duplicating the output of the encoder to the corresponding input sides in the decoder, which cannot effectively leverage the learned feature maps in both encoder and decoder paths. This study integrates several elementary modules for effectively interaction learning or knowledge transfer among the feature representations of the main network and adaptively selects more useful information. Next, we describe the integrated modules: knowledge transfer module and multi-level context gating module.

\subsection{Knowledge transfer module}
As described above, the simple feature reuse via skip connection cannot effectively fuse the learned feature representations in both encoder and decoder paths. In addition, it is known that the feature representations with more semantic information are gradually learned based on the former ones and inherit from the already learned features including those in the encoder. However, the feature representations with more semantic information in the decoder path cannot be back transferred to the encoder side for calibrating the low-level features. This study integrates a transfer module among the corresponding blocks in the encoder and decoder paths and conducts back transfer, called as internal knowledge transfer (IKT) module. Furthermore, we also aim at investigating the already learned knowledge in the released CNN models in other vision domains for aiding our deraining network training, called as external knowledge transfer (EKT) module.

\subsubsection{IKT Module:} In the conventional feature reuse with skip connection, the simple concatenate layer is used for fusing the output: ${\bf E}_{o}^{j}$ of the $j-th$ block in the encoder and the input: ${\bf D}_{i}^{j}$ of the $j-th$ block in the decoder as the real input instead of ${\bf D}_{i}^{j}$. To effectively transfer the learned semantic features in the decoder path for calibrating those in the encoder path, we consider the encoder’s output: ${\bf E}_{o}^{j}$ and the decoder’s input: ${\bf D}_{i}^{j}$ as a time sequence [${\bf E}_{o}^{j}$, ${\bf D}_{i}^{j}$] with two-time points, and employ a backward ConvLSTM ($BW\_ ConvLSTM(\cdot)$ ) unit for learning more effective features from both paths, which can be formulated as:
\begin{equation}
{\bf \hat{D}}_{i}^{j}=BW\_ConvLSTM([{\bf E}_{o}^{j}, {\bf D}_{i}^{j}])
\end{equation}
where ${\bf D}_{i}^{j}$ with semantic information is firstly inputted and the generated state in ConvLSTM calibrate the final output of ${\bf E}_{o}^{j}$ with more detailed structure. The output: ${\bf \hat{D}}_{i}^{j}$ of $BW\_ConvLSTM(\cdot)$ is adopted as the input of the $j-th$ block in the decoder. With the ConvLSTM unit, it is prospected that the learned semantic features (knowledge) in the decoder side can be effectively transferred back to the encoder side, and results in high-level representative features inside the deraining network.

\subsubsection{EKT Module:}Most existing deep learning-based deraining methods usually train the network from a randomly initialized state, and cannot leverage knowledge of the pre-trained CNN models in other vision tasks such as image classification. However, to train a good generalization model with a huge amount of unknown parameters, a large dataset is necessary, which is very tough to gather a vast number of labeled data especially for the deraining scenario. This inspires the knowledge exploiting of a pre-trained CNN to a specific under-studying task, generally called as transfer learning, which requires the similar network structure for the specific task with the pre-trained CNN model. This study attempts to explore the knowledge in the pre-trained VGG family with Imagenet dataset to overcome the isolated learning paradigm for boosting the deraining performance. Although our encoder-decoder network has different architecture with the pre-trained VGG-Net models, there are partially same structures of the encoder path in our deraining network with the shallow layers in VGG-Net. Thus, we simply transfer the parameters of the pre-trained VGG-Net's shallow layers to the first 3 blocks of the encoder path while randomly initialize the remainder structure's parameters. Then we re-train the network inheriting somewhat knowledge from the pre-trained VGG models for adapting to the new deraining task.  

\subsection{Multi-level Context Gating Module}
It is obvious that the network can obtain large amount of feature representations in different layers of multi-level blocks. However, not all learned features equivalently contribute to the subsequent extraction of high representative features and the final prediction. Recently, attention mechanism has been popularly explored to adaptively concentrate the more discriminated and effective features in the network training procedure. For example, motivated by the fact that different regions in the input may have various contributions to the final prediction such as in image classification, detection and segmentation, many work exploit the spatial attention via mutual enhancement with spatial correlation, and manifest significant improvement. Our goal aims at predicting all underlying pixel values from the rain-degraded input, and all regions should be indispensable for estimating the precise pixel values nearby. Thus, this study instead investigates the channel attention for dedicating to emphasize the channel with the underlying scene information. We suppose that the feature maps extracted by various convolutional kernels may correspond to some underlying scene layers or a part of rain layers, and propose to exploit explicit relationship between channels of the convolutional layers for gating context. We implement the context gating module via adaptively assigning a weight for each channel (channel attention) and then encoding the inputted raw feature maps. 

\begin{figure}[h]
\centering
\includegraphics[width=80mm]{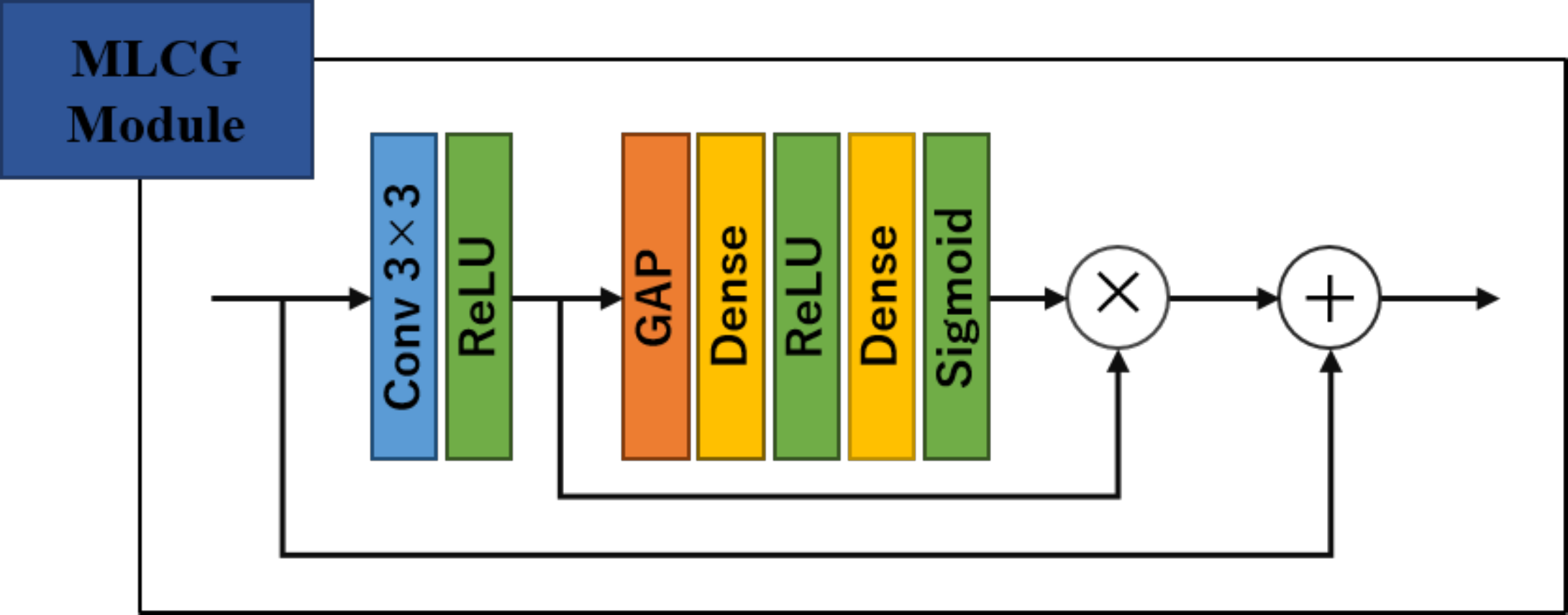}
\caption{The multi-level context gating module for MCGKT-Net.}
\label{MLCG_Module}
\end{figure}

The multi-level context gating (MLCG) module shown in Fig. \ref{MLCG_Module}, consists of two-part: squeeze and excitation (also called SE-block), and is employed on multi-level integrated feature maps [${\bf \hat{D}}_{i}^{1}$, ${\bf \hat{D}}_{i}^{2}$ , $\cdots$, ${\bf \hat{D}}_{i}^{L}$] of the encoder and decoder paths, which are also the input of the decoder’s blocks. The input feature maps to MLCG module are aggregated to generate channel contribution index by employing global average pooling (GAP) of the whole context of channels. Let's denote the input feature map to the $l-th$ MLCG module as ${\bf \hat{D}}_{i}^{l}=[{\bf \hat{d}}_{i,1}^{l}, {\bf \hat{d}}_{i,2}^{l}$, $\cdots$, ${\bf \hat{d}}_{i,C}^{l}]$, where ${\bf \hat{d}}_{i,c}^{l} \in \mathbb{R}^{W\times H}$, and the spatial squeeze (GAP) is formulated as:
\begin{equation}
s_c=f_{sq}({\bf \hat{d}}_{i,c}^{l})=\frac{1}{W\times H}\sum_h^H \sum_w^W {\bf \hat{d}}_{i,c}^{l}(h,w)
\end{equation}
where $ f_{sq}(\cdot)$ is the spatial squeeze function for compressing each two-dimensional feature map as a contribution index $ s_c$, and $ W\times H$ is the size of the $c-th$ channel feature map. Then, we employ excitation functions for capturing the channel-wise dependencies and non-linear interaction based on the global channel information ${\bf s}=[s_1, s_2, \cdots, s_C]$, which is implemented with two fully connected (FC) layers in the MLCG module. The first FC layer encodes the channel global vector ${\bf s}$ to a dimension-reduced vector with reduction ratio $r$, and the second FC layer encodes it back again to the dimension $C$ as an excitation vector, which can be expressed:
\begin{equation}
z=f_{ex} ({\bf s}, {\bf W})=\delta({\bf W}_2 \sigma({\bf W}_1 {\bf s}))
\end{equation}
where ${\bf W}_1 \in \mathbb{R}^{\frac{C}{r}\times C}$ and $ {\bf W}_2 \in \mathbb{R}^{ C \times \frac{C}{r} }$ are the parameters of the first and second FC layers, respectively, $\sigma(\cdot)$ and $\delta(\cdot)$ refer ReLU and sigmoid activation functions.
The final output of the MLCG module is generated as:
\begin{equation}
{\bf \tilde{d}}_{i,c}^{l}=f_{scale}({\bf \hat{d}}_{i,c}^{l}, z_c) = z_c {\bf \hat{d}}_{i,c}^{l}
\end{equation}
where $ f_{scale}$ denotes a channel-wise multiplication between the channel attention index $z_c$ and the input feature map.


\section{Experimental Results}

In this section, we conduct extensive experiments to validate the effectiveness of the proposed multi-level context gating knowledge transfer network (MCGKT-Net) for single image deraining. Comprehensive ablation study is given for demonstrating the effect of different modules. 

\subsection{Experimental setting-up}

\subsubsection{Implementation Details}
We implement our MCGKT-Net using Keras with TensorFlow as backbend. In network training stage, we randomly sample image patches of size 224×224 from all images as training samples, and then train the network with epochs 500. We use Adam optimizer \cite{Adam optimizer} with default parameters and  batch size of 4. The learning rate is set as $2\times10^{-4}$ .

\subsubsection{Datasets}
\begin{figure}[h]
\centering
\includegraphics[width=0.75\linewidth]{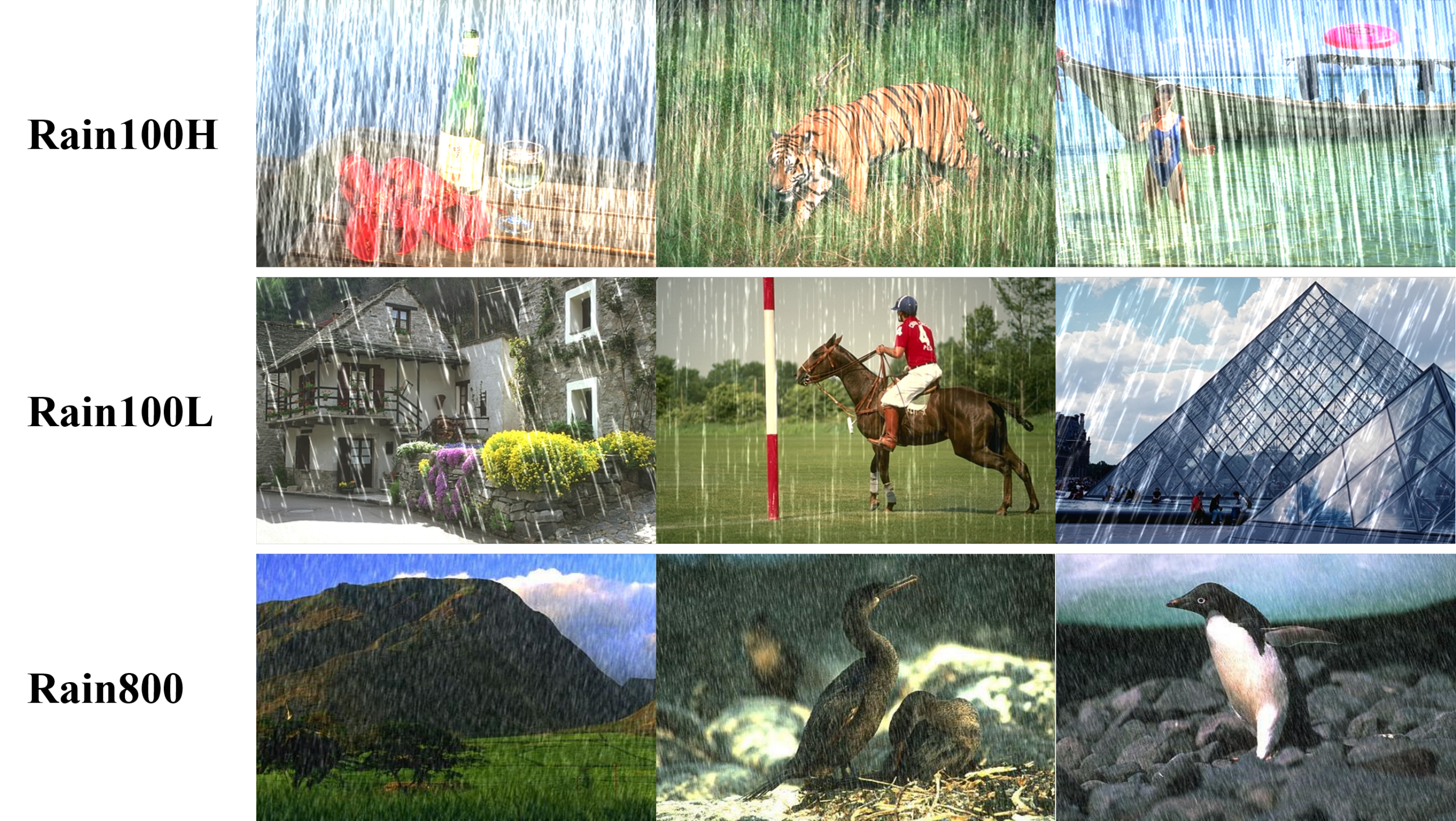}
\caption{Example rainy images from Rain100H, Rain100L and Rain800 datasets.}
\label{dataset_sample}
\end{figure}

We evaluate MCGKT-Net on three public benchmark datasets: 
\\Rain100H \cite{Rain100H_L}, Rain100L \cite{Rain100H_L}, and Rain800 \cite{Rain800}. Rain100L consists of 1800 training images and 200 test images, where the rainy images are synthesized with only one type of rain streaks while Rain100H has 1800 rainy/clean pairs as training images and 200 rainy/clean pairs as test, where the rainy images are synthesized with five directions of rain streaks. Rain800 has in total 800 images, where 700 rainy/clean pairs are as training samples and the remainders are as testing. The rainy images in Rain800 are created via adding rain streak to the clean images following the guidelines mentioned in \cite{Fu1}, which aims at generating a diverse rainy dataset via adding various intensities and orientations of rain streak to different pixels. Some synthesized rainy images from all three datasets are shown Fig. \ref{dataset_sample} manifests that the rainy images in Rain100L and Rain100H have thick and clean line structure while the rain steaks in Rain800 are much thinner and arbitrarily discontinuous without any regulation. 

\subsubsection{Evaluation Metrics}
We use two commonly metrics: i.e. peak signal to noise ratio (PSNR) and structure similarity index (SSIM~\cite{SSIM}), which measures the image structure difference and is more consistent with human perceptual measure, to evaluate the deraining performance on all three datasets.

\subsection{Comparison to the state-of-the-art methods{\vspace{-10mm}}}
\begin{figure}[H]
\centering
\includegraphics[width=0.9\linewidth]{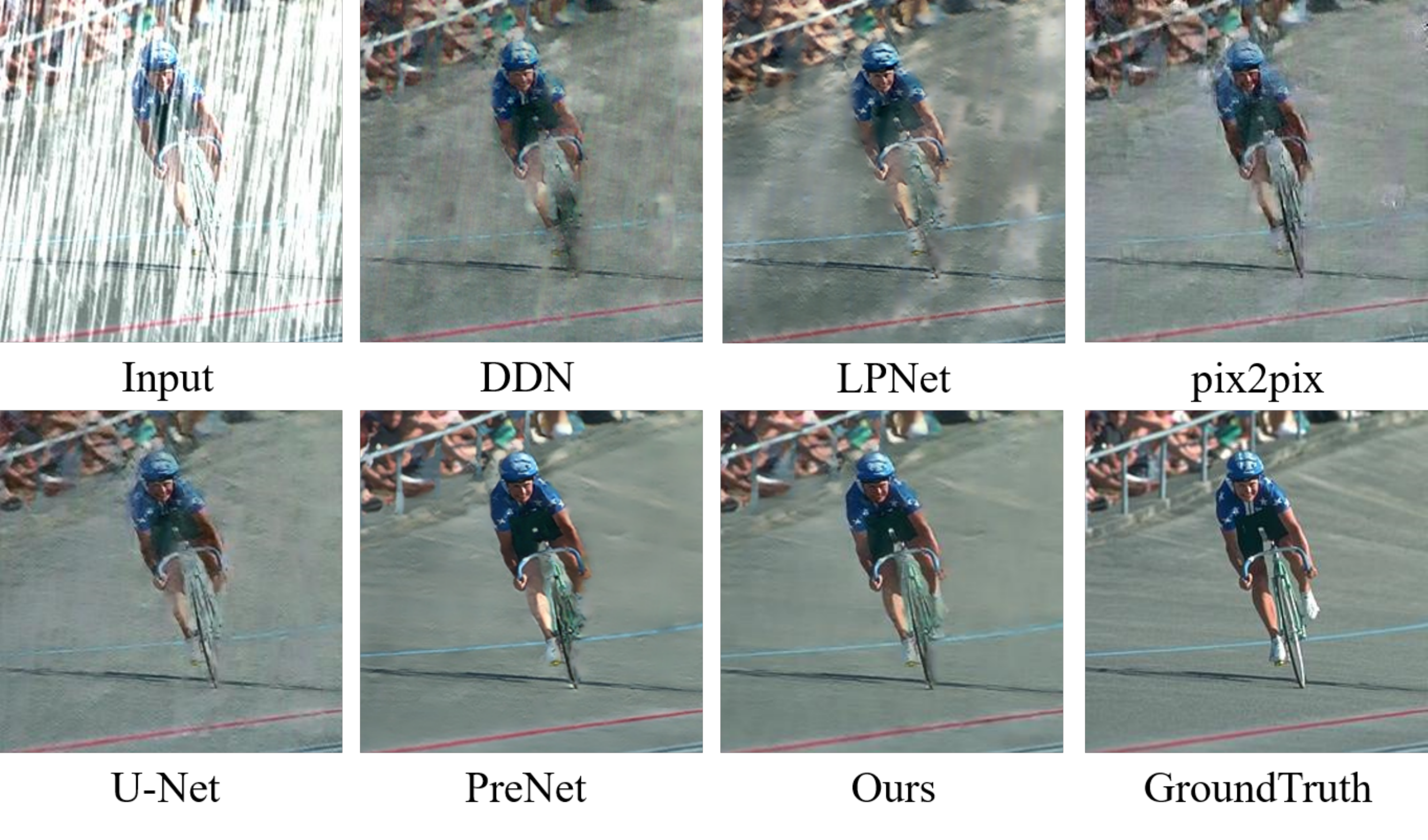}
\caption{Deraining results of different methods on Rain100H.}
\label{result_rain100H}
\end{figure}

\begin{figure}[h]
\centering
\includegraphics[width=0.9\linewidth]{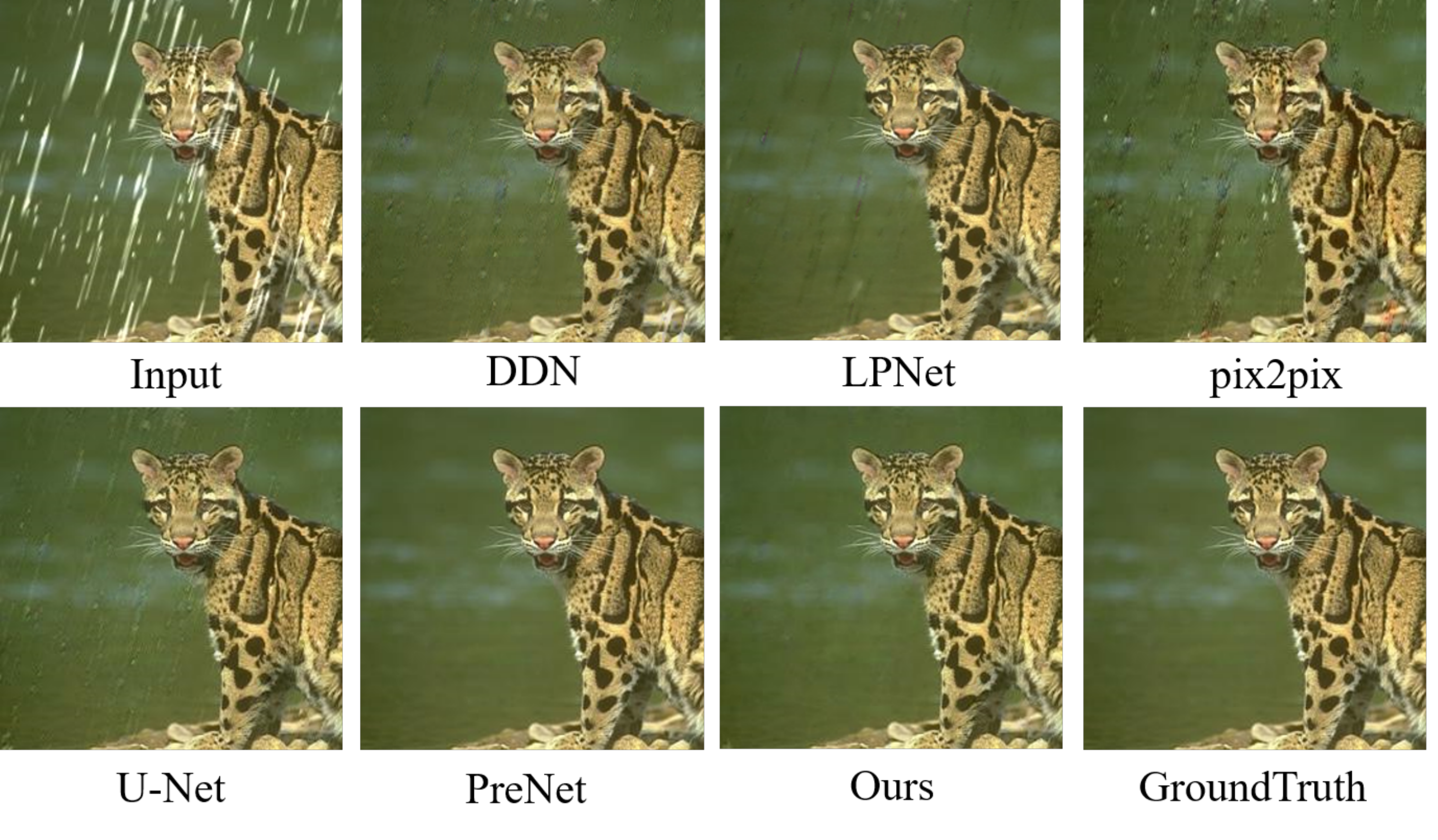}
\caption{Deraining results of different methods on Rain100L.}
\label{result_rain100L}
\end{figure}

\begin{figure*}
  \centering
  \includegraphics[width=\linewidth]{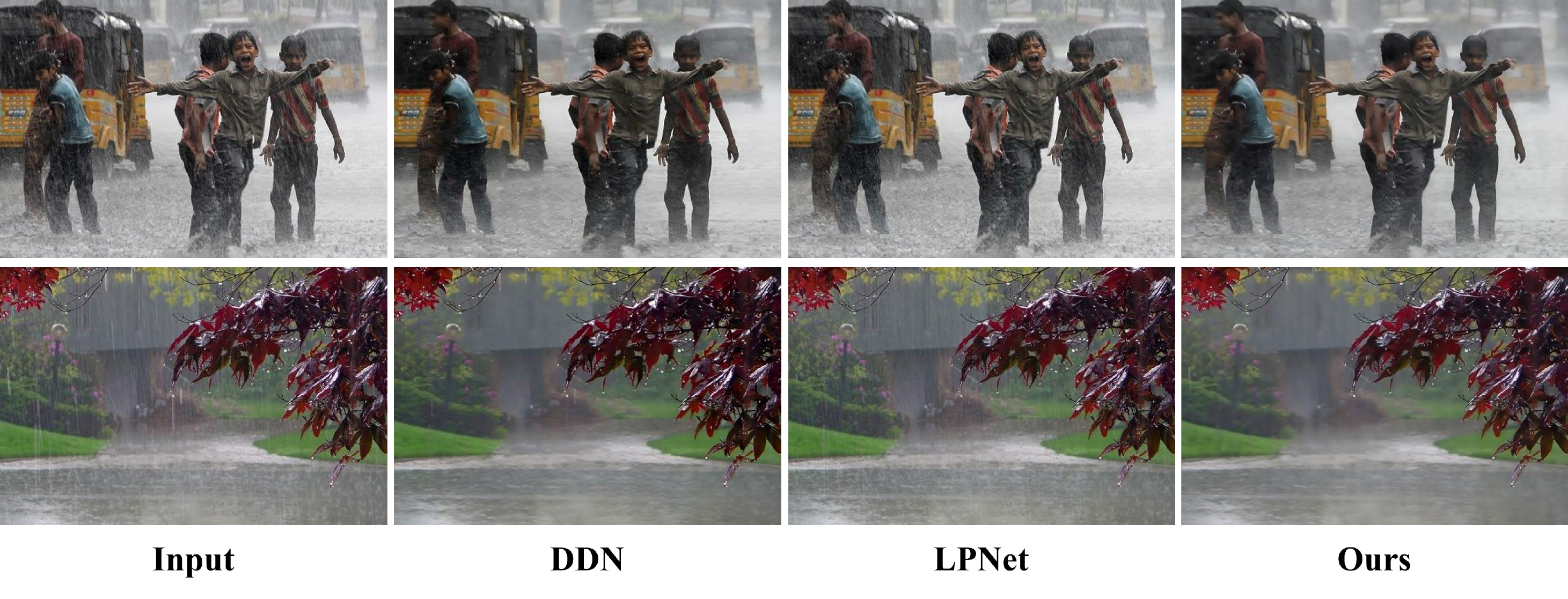}\\
 \caption{The compared results on real rainy images.}
 \label{realimage}
\end{figure*}

We compare our proposed MCGKT-Net with the state-of-the-art deraining approaches, including semi-supervised transfer learning (SEMI)~\cite{SEMI}, density-aware deraining (DIDMDN)~\cite{DIDMDN}, 
simple deep convolutional network for image SR (SRCNN)~\cite{SRCNN}, deep detail network (DDN)~\cite{DDN}, image-to-image translation (pix2pix) ~\cite{pix2pix}, lightweight pyramid network (LPNet)~\cite{LPNet}, U-Net~\cite{U-Net}, uncertainty guided multi-scale residual
learning (UMRL)~\cite{UMRL}, recurrent squeeze-and-excitation context aggregation net (RESCAN)~\cite{RESCAN}, progressive deraining network (PreNet)~\cite{PReNet}. Table \ref{aaa} shows the quantitative measure on the three datasets. From Table \ref{aaa}, we can see that our methods can outperform almost methods.
For providing visual comparison, Fig. \ref{result_rain100H} and \ref{result_rain100L} visualize the derained examples from Rain100L and Rain100H datasets using different methods, which manifests that our proposed MCGKT-Net can recover more clean images than other existing methods. Furthermore, we also provide the derained results on three real images in Fig. \ref{realimage} using our proposed MCGKT-Net and several state-of-the-art methods.

\begin{table}[H]
 \centering
 \caption{Average PSNR and SSIM  comparison on the synthetic datasets Rain100H~\cite{Rain100H_L}, Rain100L~\cite{Rain100H_L} and Rain800~\cite{Rain800}. \textcolor{red}{Red} and \textcolor{blue}{blue} colors are used to indicate top $\textcolor{red}{1^{st}}$, $\textcolor{blue}{2^{nd}}$ performance.}
 \label{aaa}
 \vspace{2mm}
 \scalebox{1.2}{
 \begin{tabular}{c|ccc}
  \hline
  Methods & Rain100H & Rain100L & Rain800 \\  \hline
  SEMI \cite{SEMI}   &       16.56/0.486     &     25.03/0.842     &   22.35/0.788    \\ 
  DIDMDN \cite{DIDMDN} & 17.35/0.524 &  25.23/0.741 & 22.56/0.818 \\
  SRCNN  \cite{SRCNN} & 18.29/0.612    & 32.63/0.936     &   25.10/0.823 \\
  DDN\cite{DDN}     & 22.08/0.788 &   31.12/0.953    &   25.10/0.823         \\ 
  pix2pix\cite{pix2pix} &  21.96/0.679  & 29.20/0.886    &      -/-       \\ 
  LPNet\cite{LPNet}    &    23.16/0.801        &      \textcolor{blue}{33.61}/\textcolor{blue}{0.958}       &       22.21/0.789         \\ 
  U-Net\cite{U-Net} &  23.28/0.741           &    30.97/0.921   &  \textcolor{blue}{26.28}/0.826    \\ 
  UMRL\cite{UMRL} & 24.91/0.810           &      31.98/0.955         & 24.37/0.819           \\
  RESCAN\cite{RESCAN} &       26.36/0.786  &    29.80/0.881        &    25.00/0.835              \\ 
  PreNet\cite{PReNet} &   \textcolor{blue}{26.77}/\textcolor{red}{0.858}   &   32.44/0.950    &  24.81/\textcolor{red}{0.865}       \\ 
  \hline
  Ours           & \textcolor{red}{27.06}/\textcolor{blue}{0.848}            &     \textcolor{red}{35.23}/\textcolor{red}{0.962}  & \textcolor{red}{27.44}/\textcolor{blue}{0.840} 
  \\ \hline

 \end{tabular}
 }
\end{table}

\subsection{Ablation studies}
\begin{figure*}
\centering
\includegraphics[width=\linewidth]{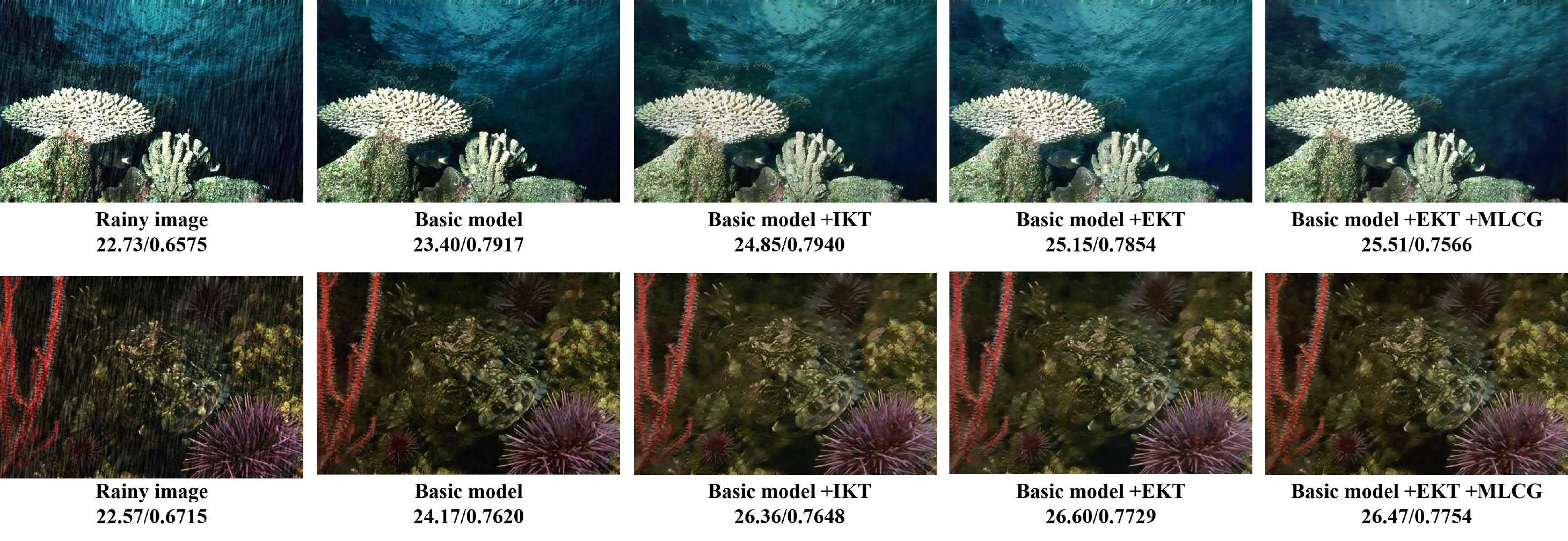}
\caption{Deraining results of ablation study on Rain800. Values below every image indicate PSNR value and SSIM \cite{SSIM} value, respectively.}
\label{rain800_result}
\end{figure*}

Our proposed MCGKT-Net is evolved from the baseline U-Net architecture, which is very simple and easy to be trained effectively. We integrated three modules: internal knowledge transfer (IKT), external knowledge transfer (EKT), and multi-level context gating (MLCG) modules into the baseline U-Net for higher representative feature learning. Therein EKT module adopted the pre-trained VGG models: VGG16 and VGG19. This section evaluates the effectiveness of the integrated modules on the baseline U-Net. Quantitative results on all three datasets are given in Table \ref{rain100hlresult}. It can be seen from Table \ref{rain100hlresult} that the proposed MCGKT-Net manifests great superiority over the baseline U-Net and its incomplete versions not integrating all proposed modules. The results of the experiment with the addition of each of the three modules in the Rain800 are shown in Fig. \ref{rain800_result}. Integration of all modules surpasses the baseline by 3.78dB, 4.26dB, 1.16dB for Rain100H, Rain100L, and Rain800, respectively.

\begin{table}[H]
 \centering
 \caption{Quantitative results by different setups on the baseline U-Net model.}
 \label{rain100hlresult}
 \vspace{2mm}
 \scalebox{0.9}{
 \begin{tabular}{c|c|cccccccc}
  \hline
  \multicolumn{2}{c|}{IKT Module} & $\times$ & \checkmark & $\times$& $\times$ & $\times$ & \checkmark & \checkmark &\checkmark \\
  \multicolumn{2}{c|}{EKT Module(VGG19)} & $\times$  &  $\times$ & \checkmark & $\times$ &$\times$ &\checkmark & $\times$ &\checkmark \\
  \multicolumn{2}{c|}{EKT Module(VGG16)} & $\times$  &  $\times$ & $\times$ & \checkmark & $\times$& $\times$ & \checkmark & $\times$ \\
  \multicolumn{2}{c|}{MLCG Module} & $\times$ & $\times$ & $\times$ & $\times$ & \checkmark & $\times$ & $\times$ & \checkmark \\
  \hline
  \hline
  \multirow{2}{*}{Rain100H}& PSNR & 23.28 & 24.61 & 26.77 &  26.45 &  24.92 & 27.03 & 26.82 & \textcolor{red}{\textbf{27.06}} \\
   &SSIM & 0.7407 &  0.7837&  0.8422 &  0.8393 &  0.7888 &0.8475 & 0.8458 & \textcolor{red}{\textbf{0.8477}}  \\ \hline

  \multirow{2}{*}{Rain100L}& PSNR & 30.97 & 33.16 &  32.04 &  31.98 & 31.10 & 35.03 & 35.02 & \textcolor{red}{\textbf{35.23}}\\
   &SSIM & 0.9210  & 0.9475 &  0.9231 &  0.9231 & 0.9131 & 0.9608 & 0.9606 & \textcolor{red}{\textbf{0.9618}}\\ \hline

  \multirow{2}{*}{Rain800}& PSNR & 26.28 & 26.76 &  26.52 &  26.38  & 25.80 & 27.32 & \textcolor{red}{\textbf{27.44}} & 27.03\\
   &SSIM & 0.8269 & 0.8307 &  0.8422 &  \textcolor{red}{\textbf{0.8428}} & 0.8190 & 0.8409 & 0.8402 & 0.8327\\ 
  \hline

 \end{tabular}
 }
\end{table}

From the results of Table \ref{rain100hlresult}, all integrated modules can improve the quantitative measures on both Rain100H and Rain100L datasets while the integration of the MLCG module on the Rain800 dataset decreases the quantitative metrics a little. From Fig. \ref{dataset_sample} as mentioned above that the rain streaks in Rain100H and Rain100L datasets exhibit some regular characteristics similar to the line structures with diverse directions while the rain streaks in the Rain800 dataset have no regular pattern most like un-regular noise. The intent of integrating the MLCG module is to adaptively emphasize and attenuate specific channels of features with some specific patterns and is expected to be oriented well to the existed rain steaks in the Rain100H and Rain100L datasets while be difficult to attenuate the noise-like rain streaks in the Rain800 dataset.


\section{Conclusion}
In this paper, we proposed a multi-level context gating knowledge transfer network for the removal of rain streaks from a single image. Taking the possible multi-layer characteristic of the rain streak in mind, we used the encoder-decoder network architecture, which itself is a multi-scale structure for feature learning, as a baseline network, and integrated several simple modules for higher representative feature learning. We employed an internal knowledge transfer module for interactively learning between the features of the encoder and decoder paths and an external knowledge transfer module for effective reuse of the knowledge preserved in a pre-trained CNN model in other task domains. Further, we explored a multi-level context gating module for adaptively emphasizing useful feature channels and attenuating the channels related to rain layers. Experimental results demonstrated that our proposed MCGKT-Net gave promising deraining performance compared with the state-of-the-art methods.


\section*{Acknowledge}
This research was supported in part by the Grant-in Aid for Scientific Research from the Japanese Ministry for Education, Science, Culture and Sports (MEXT) under the Grant No. 20K11867.

\end{document}